\documentclass[10pt,letterpaper]{article}
\usepackage{cogsci}\usepackage{amssymb}\usepackage{pslatex}\usepackage{apacite}
\usepackage{subfigure}\usepackage{times}\usepackage{helvet}\usepackage{courier}
\usepackage{url}\usepackage[T1]{fontenc}
\usepackage{color}\definecolor{mygrey}{gray}{0.80}
\usepackage{graphicx}
\newcommand{\lx}{\left}\newcommand{\rx}{\right}

\begin{document} 
\title{Spontaneous Analogy by Piggybacking on a Perceptual System}

%

\author{{\large \bf Marc Pickett} \\ NRC/NRL Postdoctoral Fellow \\ Washington, DC 20375 \\
  {\tt marc.pickett.ctr@nrl.navy.mil}
  \And {\large \bf David W.\ Aha} \\ Navy Center for Applied Research in Artificial Intelligence\\
  Naval Research Laboratory (Code 5510); Washington, DC 20375\\
  {\tt david.aha@nrl.navy.mil}}

\maketitle
\begin{abstract} 
  Most computational models of analogy assume they are given a
  delineated source domain and often a specified target domain.  These
  systems do not address how analogs can be isolated from large
  domains and spontaneously retrieved from long-term memory, a process
  we call {\em spontaneous analogy}.
  We present a system that represents relational structures as feature
  bags.  Using this representation, our system leverages perceptual
  algorithms to automatically create an ontology of relational
  structures and to efficiently retrieve analogs for new relational
  structures from long-term memory.
  %
  %
  %
  We provide a demonstration of our approach that takes a set of
  unsegmented stories, constructs an ontology of analogical schemas
  (corresponding to plot devices), and uses this ontology to
  efficiently find analogs within new stories, yielding significant
  time-savings over linear analog retrieval at a small accuracy cost.
  %
  %
  %
\end{abstract} 

\section{Spontaneous Analogy} 


In our day-to-day experience, we often generate analogies
spontaneously \cite{wharton1996remote, clement1987generation}.
That is, with no explicit prodding, we conjure up analogs to
aspects of our current situation.
For example, while reading a story, we may recognize a plot device
that is analogous to one used in another story that we read long ago.
The shared plot device may be a small part of each story, it is
usually not explicitly delineated for us or presented in isolation
from the rest of the story, and we may recognize the analogy of the
plot device even if the general plots of the two stories are not
analogous.  Somehow, we {\em segment} out the plot device and {\em
  retrieve} the analog\footnote{In our terminology, an {\em analog} is
  a substructure of a domain that is structurally similar to a
  substructure of another domain, and an {\em analogical schema} is a
  generalization of an analog.  For example, an input domain might be
  the entire story of {\em Romeo \& Juliet}, an analog would be the
  part of the story where Romeo kills Tybalt, who killed Romeo's
  friend, Mercutio (like in {\em Hamlet} where Hamlet kills Claudius,
  who killed Hamlet's father), and an analogical schema would be the
  generalized plot device of a ``revenge killing''.}  from another
story in long-dormant memory.
{\em Spontaneous analogy} is the process of efficiently retrieving an
analog from long-term memory given an unsegmented source domain such
that part of the source shares structural similarity with the analog,
though they might not share surface similarity.  This process differs
from standard models of analogy, which are given a {\em delineated}
source concept, and often a target concept.  Given a pair of analogs,
analogical mapping is relatively straightforward.  The more difficult
problem is finding the analogs to begin with.  As
\citeA{DBLP:journals/jetai/ChalmersFH92} argue ``when the program's
discovery of the correspondences between the two situations is a
direct result of its being explicitly given the appropriate structures
to work with, its victory in finding the analogy becomes somewhat
hollow''.


\begin{figure}[ht] 
  \centering{
    \subfigure[Mapping]{
      \includegraphics[width=28mm]{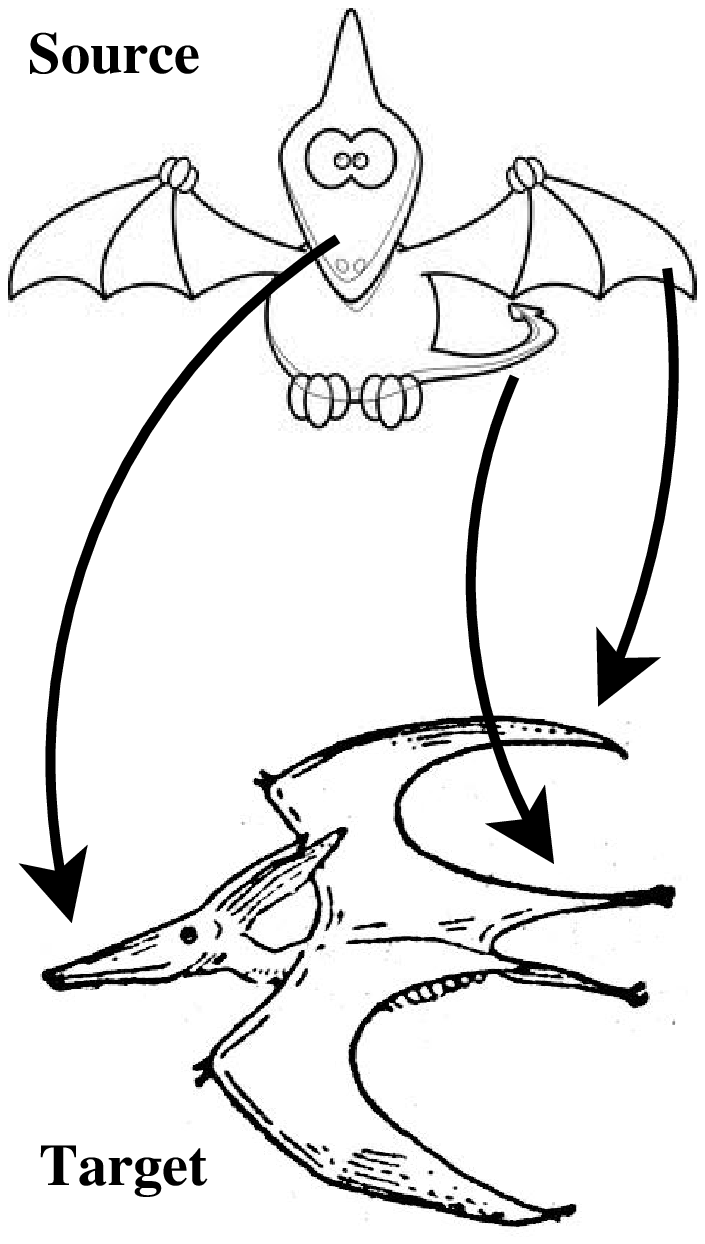}
      \label{figure:analogicalMapping}}
    \hspace{10mm}
    \subfigure[Spontaneous Retrieval]{
      \includegraphics[width=35mm]{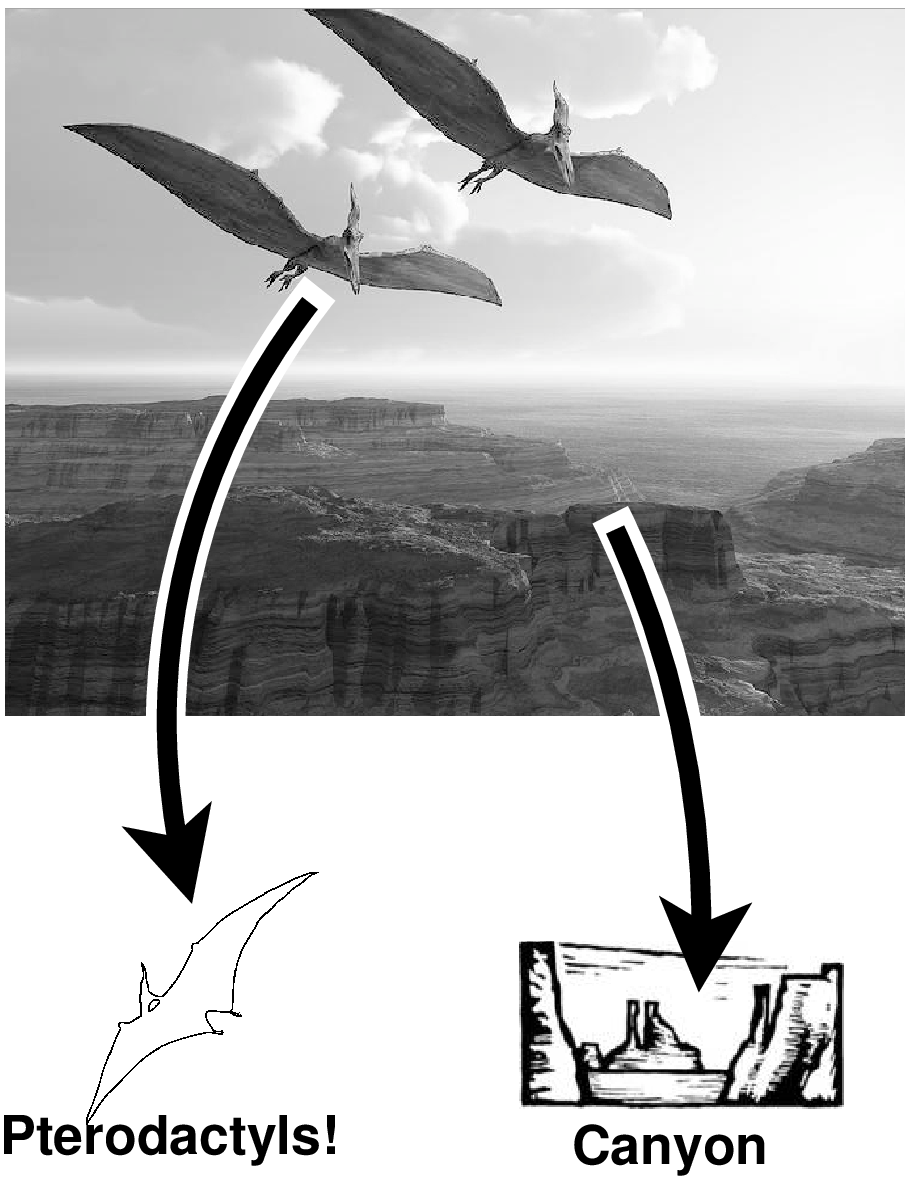}
      \label{figure:spontaneousAnalogy}}}
  \caption{{\bf An analog of Analogical Mapping vs.\ Spontaneous
      Analogy.}  In Analogical Mapping
    \subref{figure:analogicalMapping}, we are given an explicit source
    and target, free from interfering context.  In spontaneous analogy
    \subref{figure:spontaneousAnalogy}, the analogs are spontaneously
    retrieved from long-term memory.}
  \label{figure:pterodactyl}
\end{figure} 


The process of spontaneous analogy shares some properties with
low-level perception, as exemplified in Figure
\ref{figure:pterodactyl}.  Within seconds of being presented with a
visual image of a pterodactyl flying over a canyon, one can typically
describe the image using the word ``pterodactyl'', even if one has had
no special explicit recent priming for this concept, indeed even if
one has not consciously thought about pterodactyls for several years.
For us to produce the word ``pterodactyl'', we must {\em segment} the
pterodactyl from the canyon and retrieve the ``pterodactyl'' concept
from the thousands of concepts stored in memory.  We must have learned
the ``pterodactyl'' concept to begin with from unsegmented images.
This perceptual process is robust to noise: The pterodactyl in the
image could be partially occluded, ill-lit, oddly colored, or even
drawn as a cartoon, and we are still able to correctly identify this
shape (to a certain point).  Likewise, many details of the plot
devices from the above story example could be altered or obfuscated,
but this analogy would degrade gracefully.  


%



Our primary technical contribution in this paper is an algorithm
called {\em Spontol}\footnote{{\em Spontol} is short for ``{\bf
    spont}aneous analogy using the {\bf Ontol} ontology learning and
  inference algorithm''.}  that solves the problem of spontaneous
analogy: efficient parsing, storage, and retrieval of analogs from
long-term memory.  That is, given a corpus of many large unsegmented
relational structures, Spontol discovers analogical schemas that are
useful for characterizing the corpus and efficiently retrieves analogs
given a new structure.  E.g., given a set of narratives in predicate
form, Spontol discovers plot devices and analogs between the stories.
We know of no prior work that scales to this task when the number of
narratives and statements per narrative are both in the hundreds.

In the remainder of this paper, we describe related work (Section
\ref{section:related}), give background on {\em perceptual systems}
(Section \ref{section:perceptualMethods}), describe the Spontol
algorithm, which transforms the problem of spontaneous analogy into a
``perceptual'' problem (Section \ref{section:analogyAsPerception}),
demonstrate Spontol's performance on a story database (Section
\ref{section:demonstration}), discuss implications and shortcomings of
Spontol, and conclude (Section \ref{section:conclusion}).

\section{Related Work} 
\label{section:related}


There has been earlier work on the problem of analogy in the absence
of explicitly segmented domains.  The COWARD system
\cite{baldwin+goldstone:2007} addresses this problem by searching for
mappings within a large graph, essentially searching for isomorphic
subgraphs.
SUBDUE \cite{holder+cook+djoko:1994} compresses large graphs by
breaking them into repeated subgraphs, but is limited in that its
output must be a strict hierarchy, and would be unable to discover the
lattice structure of the concepts in Figure \ref{figure:zooUpper}.
Nauty \cite{mckay:1981} uses a number of heuristics to efficiently
determine whether one graph is a subgraph of another, but this must be
given source and target graphs to begin with.
We can also apply The Chunker (described in Section
\ref{section:perceptualMethods}) to feature bag graphlet kernels
\cite{SheVisPet2009}, which are related to Spontol's transform $T$ in
that both represent partial graphs, but this earlier work applies only
for cases where there is one kind of entity, one kind of relation, and
only binary relations, while Spontol works for multiple kinds of
entities and relations, including relations of large arity.


The MAC phase of MAC/FAC \cite{forbus1995mac} bears some relation to
our spontaneous analog retrieval.
MAC uses vectors of content, such as the number of nodes and edges in
a graph, as a heuristic for analog retrieval.
However, in cases where the subgraph in question is a part of a much
larger graph, the heuristics that MAC uses are drowned out by the
larger graph.
Likewise, ARCS \cite{thagard1990analog} also assumes that analogs have
been delineated (i.e., it matches an entire source domain, rather than
a substructure).
{\em SEQL} \cite{Kuehne00seql:category} generalizes relational
concepts, but doesn't build a hierarchical ontology of analogical
schemas.

%


There has been some work on representing structures as feature
vectors.  For example, Holographic Reduced Representations
have been used to implement Vector Symbolic Architectures in which
there is a correlation between vector overlap and structural
similarity \cite{gayler2009distributed}.  This work is limited in that
it requires vectors of length 10,000 to represent very small graphs
($\leq$ 10 nodes), and only represents binary relations of a single
type, so this approach is not directly extendable to relational
structures such as the stories in our demonstration.  This is also a
limitation for the system proposed by \citeA{rachkovskij2012building}.
Both these systems are also limited in that they are unable to exploit
partial analogical schemas.  That is, a partial overlap in these
systems' vectors does not correspond to a common subgraph in the
corresponding structures.  These systems stand in contrast to Spontol,
which is able to represent larger structures and efficiently find
common substructures.

\section{Background: Perceptual Systems} 
\label{section:perceptualMethods}



Spontol transforms relational structures into feature bags so that
their surface similarity corresponds to the structural similarity of
the relational structures.  After Spontol has made this
transformation, the problem of spontaneous analogy is reduced to the
problem of feature overlap, and any of several existing ``perceptual''
systems can be used to find and exploit patterns in feature vectors.
Our implementation of Spontol uses a model inspired by the human
sensory cortices (auditory, visual, tactile) called {\em Ontol}
\cite{pickett:2011}.  Ontol is a pair of algorithms, both of which are
given ``sensor'' inputs (fixed-length, real-valued non-negative
vectors).  The first algorithm constructs an ontology that concisely
encodes the inputs.  For example, given a set of vectors representing
visual windows from natural images, Ontol produces a feature hierarchy
loosely modeled on that seen in the visual cortex.
The second algorithm takes as input an ontology (produced by the first
algorithm) and a new vector, and {\em parses} the vector.  That is, it
produces as output the new vector encoded in the higher-level features
of the ontology.  In addition to ``bottom-up'' parsing, the second
algorithm also makes ``top-down'' predictions about any unspecified
values in the vector.

Ontol is ignorant of the modality of its input.  That is, Ontol is
given no information about what sensory organ is producing its inputs.
Because of this ignorance, we are able to leverage Ontol to find
patterns in abstract ``sensory'' inputs that are actually encodings of
relational structures.  


\subsection{Ontology Learning} 

Ontol's ontology formation algorithm, called {\em The Chunker}, seeks
to find concepts (or {\em chunks}) that allow for concise
characterization of vectors.  Since chunks themselves are vectors, The
Chunker is applied recursively to create an ontology.  In essence,
this algorithm is similar to the {\em recursive block pursuit}
algorithm described by \citeA{si2011unsupervised} in that both search
for large frequently occurring sets of features.  The Chunker differs
in that it allows for multiple inheritance, while recursive block
pursuit creates only strict tree structures.  In Section
\ref{section:analogyAsPerception}, we show the importance of this
property for finding multiple analogical schemas within a single
relational structure.
For simplicity, we describe the discrete binary version of The Chunker
algorithm ({\sf chunk}$\lx(B\rx)$, which takes as input a set $B$ of
feature bags and produces an ontology $\Omega$) provided by
\citeA{pickett:2011}, but this can be modified for continuous vectors.
In this version, each vector is treated as a set, with a value of $1$
for feature $f$ signifying inclusion of $f$ in the set, and a value of
$0$ signifying exclusion.

\begin{figure}[h] 
  \centering\includegraphics[width=80mm]{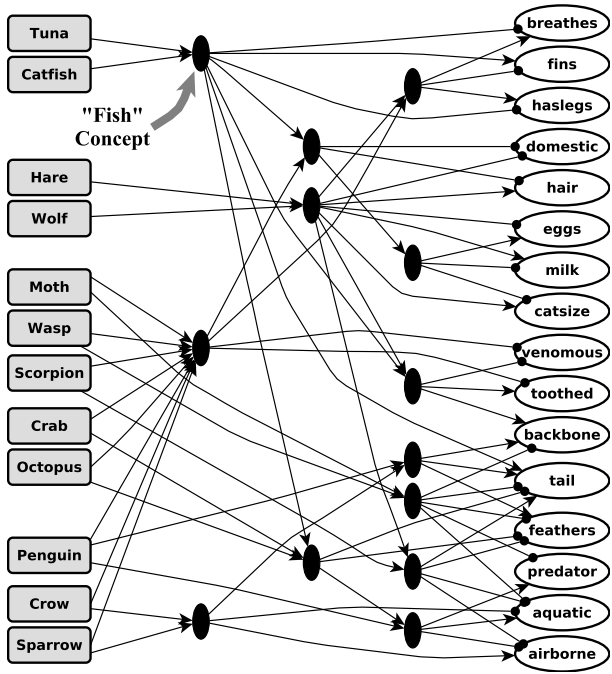}
  \caption{{\bf The Zoo Ontology with some instances.}  Instances are
    individual animals shown on the left, and base features are on the
    right.  Black nodes in the middle correspond to higher-level
    features.  The concept that corresponds to ``fish'' is marked.
    Inhibitory links are shown as dark circles.}
  \label{figure:zooUpper}
\end{figure}

The Chunker searches for intersections among existing feature bags and
proposes these as candidates for new concepts.  Each candidate is
evaluated by how much it would compress the ontology, then the best
candidate is selected and added to the set of feature bags, and the
process is repeated until no candidates are found that further reduce
the description length of the ontology.
Figure \ref{figure:zooUpper} shows the ontology constructed by this
algorithm when applied to an animal dataset, where the ``sensory
percepts'' are features for each animal\footnote{A full description
  and implementation of The Chunker, as well as source code for our
  demonstration of Spontol can be downloaded at
  \url{http://marcpickett.com/src/analogyDemo.tgz}.
  \label{footnote:URL}}.

\subsection{Parsing and Prediction} 

Given an ontology and a new instance, Ontol's {\sf parse}$\lx(b,
\Omega\rx)$ algorithm characterizes the feature bag instance $b$ using
the higher-level features in the ontology $\Omega$.  For example,
given a new animal (a goldfish) that doesn't breathe, has fins, has no
feathers, and is domestic, Ontol will parse the animal as an instance
of the {\em fish} concept, with the exception that it is domestic.  If
Ontol is given no other information about the animal, it will also
perform top-down inference, and {\em unfold} the fish concept to
predict that the new instance has eggs, no hair, has a tail, etc..
This latter step is called ``top-down prediction''.  Ontol searches
for the parse that minimizes the description length of the instance.
In our goldfish example, the ``raw'' description of the goldfish
consists of 4 elements, while the ``compressed'' description has only
2 elements.

Although the parsing problem is NP-complete, a single bottom-up pass
can be performed in logarithmic time \cite{pickett:2011}.
Importantly, Ontol examines only a small subset of the concepts and
instances while parsing.  This means that, when judging concept
similarity, Ontol does not need to compare each of its $n$ nodes.
This property is important for spontaneous analog retrieval (described
below).

\section{Analogy as Perception} 
\label{section:analogyAsPerception}

We now describe a method for transforming relational structures into
sparse feature vectors (or feature bags) such that the problem of
analog retrieval is reduced to the problem of percept parsing.  An
example of this process is shown for the {\em Sour Grapes} fable in
Figure \ref{figure:sourgrapesbig}.
%
%
For this process, we rely on a transform $T$ (described below) that
takes a small relational structure and converts it into a feature bag
(exemplified in Figure \ref{figure:sourgrapesTransform}).  The size of
relational structure is limited for $T$ because $T$'s runtime is
quadratic in the size of the structure.
We view this limitation as acceptable because people generally cannot
keep all the details of an entire lengthy novel (or all the workings
of a car engine) in working memory.  Generally, people focus on some
aspect of the novel, or some abstracted summary of the novel (or engine).
Therefore, we break each large relational structure into multiple
overlapping {\em windows}.  A window is a small set of connected
statements, where two statements are connected if they share at least
one argument.
Spontol exploits a principle akin to one used by the HMax model of the
visual cortex \cite{riesenhuber+poggio:1999}: as the number of windows
for a relational structure increases, the probability decreases that
another structure has the same windows without being isomorphic to the
first.

The process for building an ontology of analogical schemas from large
relational structures, called {\sf Spontol-Build}, is described in
Figure \ref{figure:spontol1}.  This algorithm extracts $numWindows$
windows from each large relational structure and transforms them into
feature bags (exemplified in Figure \ref{figure:sourgrapesWindows})
and chunks these feature bags to create an ontology of windows called
$windowOntology$.  {\sf Spontol-Build} then re-encodes the windows by
parsing them using this ontology, and re-encodes the larger structures
(from which the windows came) as a feature bag of the parsed windows.
Finally, {\sf Spontol-Build} runs another pass of chunking on the
re-encoded structures to generate the schema ontology.

\begin{figure}[ht] 
  \caption{{\bf Spontol's Ontology Learning Algorithm}}
  \label{figure:spontol1}
  \fbox{\footnotesize
    \begin{minipage}[b]{0.95\linewidth} 
      \begin{tabbing} 
        \hspace{10pt}\=\hspace{10pt}\=\hspace{10pt}\=\hspace{10pt}\=\hspace{10pt}\=\hspace{10pt}\=\hspace{40pt} \= \kill
        // Creates an ontology of schemas given a set of structures $S$.
        \\ // $numWindows$ is the number of windows to grab per structure.
        \\ // $windowSize$ is the number of statements per window.
        \\ {\bf define} {\sf Spontol-Build} $\lx(S, numWindows, windowSize\rx)$
        \\\> // Randomly grab windows from each structure,
        \\\> // and transform them into feature bag form.
        \\\> {\bf foreach} $s \in S$ {\bf ;} {\bf for} $i=1, \cdots, numWindows$
        \\\>\> {\bf let} $w_{s,i}$ = {\sf grabConnectedStatements} $\lx(s, windowSize\rx)$
        \\\>\> {\bf add} $T\lx(w_{s,i}\rx)$ {\bf to} $allWindows$
        \\\> // Run The Chunker to generate the window ontology
        \\\> $windowOntology$ = {\sf chunk} $\lx(allWindows\rx)$
        \\\> // Re-encode each structure using the reduced-size windows.
        \\\> {\bf foreach} $s \in S$ {\bf ;} {\bf for} $i=1, \cdots, numWindows$
        \\\>\> {\bf add} {\sf parse} $\lx(T\lx(w_{s,i}\rx), windowOntology\rx)$ {\bf to} $bigWindows_s$
        \\\> // Run The Chunker to generate the schema ontology.
        \\\> $schemaOntology$ = {\sf chunk} $\lx(bigWindows\rx)$
        \\\> {\bf return} $schemaOntology$, $windowOntology$
      \end{tabbing} 
    \end{minipage} 
  }
\end{figure} 

The process of spontaneous analog retrieval, called {\sf
  Spontol-Retrieve}, is given in Figure \ref{figure:spontol2}.  When
given a new relational structure $s$, we encode $s$ by extracting
windows from it, parsing these using the $windowOntology$, then
parsing the feature bag representation using the $schemaOntology$.
This yields a set of schemas that are contained in $s$.

\begin{figure}[ht] 
  \caption{{\bf Spontol's Spontaneous Analogy Algorithm}}
  \label{figure:spontol2}
  \fbox{\footnotesize
    \begin{minipage}[b]{0.95\linewidth} 
      \begin{tabbing} 
        \hspace{10pt}\=\hspace{10pt}\=\hspace{10pt}\=\hspace{10pt}\=\hspace{10pt}\=\hspace{10pt}\=\hspace{110pt} \= \kill
        // Finds analogical schemas for relational structure $s$.
        \\ // $schemaOntology$ is the schema ontology.
        \\ // $windowOntology$ is the window ontology.
        \\ // $numWindows$ is the number of windows to grab per structure.
        \\ // $windowSize$ is the number of statements per window.
        \\ {\bf define} {\sf Spontol-Retrieve} $\lx(s, \cdots, windowSize\rx)$\>\>
        \\\> // Randomly grab windows from $s$, 
        \\\> // transform them into feature bag form,
        \\\> // and parse them using the window ontology.
        \\\> {\bf for} $i=1, \cdots, numWindows$
        \\\>\> $w_{i}$ = {\sf grabConnectedStatements} $\lx(s, windowSize\rx)$
        \\\>\> {\bf add} {\sf parse} $\lx(T\lx(w_{i}\rx), windowOntology\rx)$ {\bf to} $bag_s$
        \\\> // Parse $bag_s$, the bag representation of $s$
        \\\> $relevantSchemas$ = {\sf parse} $\lx(bag_s, schemaOntology\rx)$
        \\\> {\bf return} $relevantSchemas$
      \end{tabbing} 
    \end{minipage} 
  }
\end{figure} 

\subsection{Transforming Small Relational Structures} 

Here, we describe an operation $T$, which transforms a (small)
relational structure into a feature bag.  In our demonstration, we
assume that the relational structure is described in predicate logic,
but our approach is not limited to this representation.  We consider a
relational structure to be a {\em set} of relational statements, where
each statement is either a relation (of fixed arity) with its
arguments, or the special relation {\tt sameAs}, which uses the syntax
{\tt sameAs <name> (<relation> <arg1> <arg2> ...)}.  The {\tt sameAs}
relation allows for statements about statements.  E.g., the statements
in Figure \ref{figure:sourgrapes} encode (among other things) that ``a
fox {\em decides that} the grapes are sour''.

\begin{figure} 
  \centering
  \subfigure[English (for clarity)]{
    \label{figure:sourgrapesenglish}
    \fbox{\footnotesize
      \begin{minipage}[b]{0.95\linewidth} 
        ``A fox wanted some grapes, but could not get them.  This
        caused him to decide that the grapes were sour, though the
        grapes weren't.  Likewise, men often blame their failures on
        their circumstances, when the real reason is that they are
        incapable.''
      \end{minipage} 
    }
  }
  
  \subfigure[Predicate Form (Spontol's actual input)]{
    \label{figure:sourgrapes}
    \fbox{
      \begin{minipage}[b]{0.95\linewidth} 
        {\tiny \tt
          \begin{tabular}{@{\hspace{0pt}}l@{\hspace{5pt}}|@{\hspace{5pt}}l@{\hspace{5pt}}|@{\hspace{5pt}}l@{\hspace{0pt}}} 
            fox Of3Fox       &cause m34 m33            &sameAs f36 (sour Of3Grapes)
            \\ false f36        &grapes Of3Grapes         &sameAs f35 (decide Of3Fox f36)
            \\ cause f34 f35    &incapable Of3Men         &sameAs f34 (get Of3Fox Of3Grapes)
            \\ false f34        &decide Of3Fox f36        &sameAs m34 (incapable Of3Men)
            \\ men Of3Men       &sameAs m33 (fail Of3Men) &blameFor Of3Men concCircum m33
            \\ fail Of3Men      &want Of3Fox Of3Grapes    &circumstances concCircum
          \end{tabular} 
          
        }
      \end{minipage} 
    }
  }
  
  \subfigure[Transforming a Window]{
    \label{figure:sourgrapesTransform}
    \centering{\tiny
      \begin{tabular}{@{\hspace{-5pt}}l@{\hspace{-5pt}}c@{\hspace{-5pt}}r@{\hspace{-7pt}}} 
        \fbox{\tiny\tt
          \begin{tabular}{@{\hspace{0pt}}l@{\hspace{-3pt}}} 
            blameFor Of3Men concCircum m33
            \\ sameAs m33 (fail Of3Men)
            \\ fail Of3Men
            \\ circumstances concCircum
            \\ men Of3Men
            \\ incapable Of3Men
          \end{tabular} 
        }
        &
        \begin{tabular}{c} 
          {\large $T$}
          \\ {\Huge\bf$\Rightarrow$}
        \end{tabular} 
        &
        \colorbox{mygrey}{\tiny\tt
          \begin{tabular}{@{\hspace{0pt}}l@{\hspace{-3pt}}} 
            blameFor1=blameFor3.fail1
            \\ circumstances1=blameFor2
            \\ fail1=blameFor3.fail1
            \\ fail1=blameFor1
            \\ incapable1=blameFor3.fail1
            \\ incapable1=blameFor1
            \\ incapable1=fail1
            \\ men1=blameFor3.fail1
            \\ men1=blameFor1
            \\ men1=fail1
            \\ men1=incapable1
          \end{tabular} 
        }
      \end{tabular} 
    }
  }
  
  \subfigure[Many Transformed Windows]{
    \label{figure:sourgrapesWindows}
    \centering{
      \fbox{\tiny
        \begin{tabular}{@{\hspace{-5pt}}l@{\hspace{-10pt}}r@{\hspace{-7pt}}} 
          \begin{tabular}{c} 
            \colorbox{mygrey}{\tiny\tt
              \begin{tabular}{@{\hspace{0pt}}l@{\hspace{-3pt}}} 
                blameFor1=blameFor3.fail1
                \\ circumstances1=blameFor2
                \\ fail1=blameFor3.fail1
                \\ fail1=blameFor1
                \\ incapable1=blameFor3.fail1
                \\ incapable1=blameFor1
                \\ incapable1=fail1
                \\ men1=blameFor3.fail1
                \\ men1=blameFor1
                \\ men1=fail1
                \\ men1=incapable1
              \end{tabular} 
            }
            \\
            \\
            \colorbox{mygrey}{\tiny\tt
              \begin{tabular}{@{\hspace{0pt}}l@{\hspace{-3pt}}} 
                false1.sour1=decide2.sour1
                \\ decide1=cause2.decide1
                \\ decide2=cause2.decide2
                \\ false1=cause2.decide2
                \\ false1=decide2
              \end{tabular} 
            }
            \\
            \\
            {\Huge .}\\
            {\Huge .}\\
            {\Huge .}\\
            {\Huge .}\\
            {\Huge .}
          \end{tabular} 
          &
          \begin{tabular}{c} 
            \colorbox{mygrey}{\tiny\tt
              \begin{tabular}{@{\hspace{0pt}}l@{\hspace{-3pt}}} 
                cause2.fail1=blameFor3.fail1
                \\ blameFor1=blameFor3.fail1
                \\ blameFor1=cause2.fail1
                \\ cause2=blameFor3
                \\ fail1=blameFor3.fail1
                \\ fail1=cause2.fail1
                \\ fail1=blameFor1
                \\ men1=blameFor3.fail1
                \\ men1=cause2.fail1
                \\ men1=blameFor1
                \\ men1=fail1
              \end{tabular} 
            }
            \\
            \\
            \colorbox{mygrey}{\tiny\tt
              \begin{tabular}{@{\hspace{0pt}}l@{\hspace{-3pt}}} 
                blameFor1=blameFor3.fail1
                \\ fail1=blameFor3.fail1
                \\ fail1=blameFor1
                \\ incapable1=blameFor3.fail1
                \\ incapable1=blameFor1
                \\ incapable1=fail1
                \\ men1=blameFor3.fail1
                \\ men1=blameFor1
                \\ men1=fail1
                \\ men1=incapable1
              \end{tabular} 
            }
            \\
            {\Huge .}\\
            {\Huge .}\\
            {\Huge .}
          \end{tabular} 
        \end{tabular} 
      }
    }
  }
  \caption {{\bf Transforming the {\em Sour Grapes} Story.}  We show
    the transformation of {\em Sour Grapes} from predicate form to
    feature bag form.  For clarity, we show an English paraphrase of
    the story \subref{figure:sourgrapesenglish}, though the input to
    Spontol has already been encoded in the predicate form shown in
    \subref{figure:sourgrapes}, which shows the story as a set of 18
    statements.  In \subref{figure:sourgrapesTransform}, we show a
    window $w$ from the story and its feature bag transform
    $T\lx(w\rx)$.  Finally, the story is represented as many 
    transformed windows \subref{figure:sourgrapesWindows}.}
  \label{figure:sourgrapesbig}
\end{figure} 

Given a small relational structure $s$ ($\lesssim 10$ statements), $T$
transforms $s$ into a feature bag using a variant of conjunctive
coding.  That is, $T$ breaks each statement into a set of roles and
fillers.  For example, the statement {\tt want Of3Fox Of3Grapes} has
two roles and fillers, namely the two arguments of the {\tt want}
relation.  So $T$ breaks this statement into {\tt want1=Of3Fox} and
{\tt want2=Of3Grapes}, where {\tt want2} means the 2nd argument of
{\tt want} (i.e., the ``want{\em ed''}).  $T$ then creates one large
set of all the roles and their fillers.  If there are multiple
instances of a relation, it gives them an arbitrary lettering (e.g.,
{\tt wantB1=Of3Fox}).  $T$ makes a special case for the {\tt sameAs}
relation.  In this case, $T$ uses a {\em dot} operator to replace the
intermediate variable.  For example, the statements {\tt sameAs f35
  (decide Of3Fox f36)} and {\tt sameAs f36 (sour Of3Grapes)} would
yield {\tt decide2.sour1=Of3Grapes}.  The dot operator allows $T$ to
encode nested statements (i.e., statements about statements).
Given a set of roles and fillers, $T$ then {\em chains} the fillers to
get {\em filler equalities}.  For example, if we have that {\tt
  decide1=Of3Fox} and {\tt want1=Of3Fox}, then chaining gives us {\tt
  decide1=want1}.  Chaining is essential for recognizing structural
similarity between relational structures, and allows us to side-step a
criticism of conjunctive coding and tensor products: that the code for
{\tt wantB1=Of3Fox} may have no overlap with the code for {\tt
  want1=Of3Fox} \cite{hummel+holyoak+green+doumas:2004}.  Chaining
introduces the code for {\tt wantB1=want1}, which makes the similarity
apparent when searching for analogs (these ``chained'' features are a
core difference between MAC's content vectors and our feature bags).
After chaining the roles and fillers, $T$ treats each of these
role-filler bindings as an atomic feature.  Note that, when we treat
roles and fillers as atomic features, Ontol doesn't recognize overlap
among feature bags unless they share exactly the same feature.  For
example, the atomic feature {\tt wantB1=Of3Fox} has no more
resemblance to {\tt want1=Of3Fox} for Ontol than it does for any other
feature.  Also note that the ordering of the roles in each feature is
arbitrary but consistent ($T$ uses reverse alphabetical order), so
there is a {\tt men1=incapable1} feature, but not an {\tt
  incapable1=men1} feature.
The left side of Figure \ref{figure:sourgrapesTransform} shows a window
taken from the sour grapes story from Figure \ref{figure:sourgrapes}.
On the right side is the feature bag transform of this set of 6
statements, consisting of 11 atoms.

\section{Demonstration} 
\label{section:demonstration}

We applied Spontol to a database of 126 stories provided by
\citeA{thagard1990analog}.  These include 100 fables and 26 plays all
encoded in a predicate format, where each story is a set of unsorted
statements.  An example story in predicate form is shown in Figure
\ref{figure:sourgrapes}.  Note that although the predicates and
arguments have English names, our algorithm treats all these as
gensyms except for the special {\tt sameAs} relation.
In this encoding, the smallest story had 5 statements, while the
largest had 124 statements, with an average of 39.5 statements.  

\begin{figure}[ht] 
  \centering\includegraphics[width=80mm]{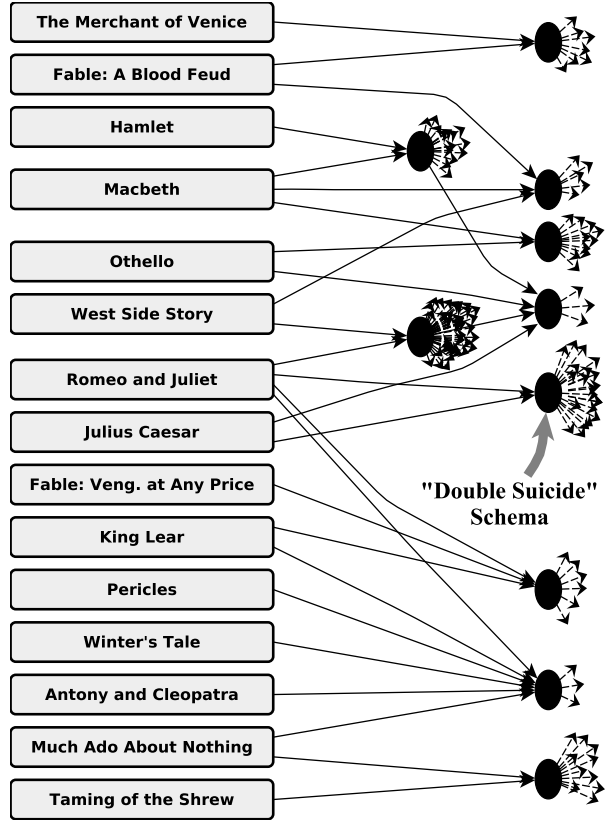}
  \caption{{\bf Part of the ontology Spontol learned from the story
      dataset.} As in the Zoo Ontology in Figure
    \ref{figure:zooUpper}, black ovals represent higher level
    concepts.  The ``raw'' features (corresponding to the white ovals
    in Figure \ref{figure:zooUpper}) are omitted due to space
    limitations.  Instead, we show the outgoing edges from each black
    oval.  While in the Zoo Ontology, the higher level concepts
    correspond to shared surface features, in this figure, high level
    concepts correspond to shared structural features, or {\em
      analogical schemas}.  For example, the highlighted oval on the
    right represents a {\em Double Suicide} schema, which happens in
    both {\em Romeo \& Juliet} and in {\em Julius Caesar}.}
  \label{figure:thagard}
\end{figure} 

We ran {\sf Spontol-Build} on these stories using $numWindows$ = $100$
and $windowSize$ = $20$ which produced an ontology of stories, part of
which is shown in Figure \ref{figure:thagard}.
In this figure we see a ``Double Suicide'' analogical schema found in
both {\em Romeo \& Juliet} and in {\em Julius Caesar}.  In the
former, Romeo thinks that Juliet is dead, which causes him to kill
himself.  Juliet, who is actually alive, finds that Romeo has died,
which causes her to kill herself.  Likewise, in {\em Julius Caesar},
Cassius kills himself after hearing of Titinius's death.  Titinius,
who is actually alive, sees Cassius's corpse, and kills himself.
The largest schema found (in terms of number of outgoing edges) was
that shared by {\em Romeo \& Juliet} and {\em West Side Story}, which
are both stories about lovers from rival groups.  The latter doesn't
inherit from the Double Suicide schema because Maria (the analog of
Juliet), doesn't die in the story, and Tony (Romeo's analog)
meets his death by murder, not suicide.
Some of the schemas found were quite general.  For example, the oval
on the lower right with 6 incoming edges and 3 outgoing edges
corresponds to the schema of ``a single event has two significant
effects''.  And the oval above the Double Suicide oval corresponds to
the schema of ``killing to avenge another killing''.

{\sf Spontol-Retrieve} uses this schema ontology to efficiently
retrieve schemas for a new story, which can be used to make inferences
about the new story in a manner analogous to the ``goldfish'' example
from Section \ref{section:perceptualMethods}.
To evaluate the efficiency of {\sf Spontol-Retrieve}, we randomly
split our story dataset into 100 training stories and 26 testing
stories.  We then used an ontology learned from the training set, and
measured the number of comparisons needed to retrieve schemas (during
{\sf parse}) for the testing set.  We compare this approach to
MAC/FAC, which, during the MAC phase, visits each of the 100 training
stories.
Whereas MAC/FAC returns entire stories, {\sf Spontol-Retrieve} returns
analogical {\em schemas} (just as a visual system would return a
generic ``pterodactyl'' concept rather than specific instances of
pterodactyls).  For comparison, we modify {\sf Spontol-Retrieve} to
return the set of instances that inherit from $relevantSchemas$,
rather than just the schemas.

\begin{table}[ht] 
  \caption{{\bf Speed/Accuracy Comparison of Spontol}}
  \label{table:results}
  \centering{
    \fbox{
      \begin{tabular}{lll} 
        & \multicolumn{1}{c}{\footnotesize \bf Accuracy}
        & \multicolumn{1}{c}{\footnotesize \bf Average \# Comparisons}
        \\ {\bf MAC/FAC}      & 100.00\% $\pm$ .00\%   & 100.00 $\pm$ .00
        \\  {\bf Spontol} & \phantom{0}95.45\% $\pm$ .62\%  &  \phantom{0}15.43 $\pm$ .20
      \end{tabular} 
    }
  }
\end{table} 

Results are shown in Table \ref{table:results}, averaged over 100
trials.  We show accuracy (and standard error) for both systems
measured as the percentage of stories correctly retrieved, where a
story was determined to be correct if it was retrieved by MAC/FAC.
Spontol effectively improves on a linear (in the number of structures)
case-by-case comparison to an ``indexed'' logarithmic-time look-up at
a slight cost of accuracy.  Therefore, Spontol requires orders of
magnitude fewer comparisons than MAC/FAC, {\em or any linear look-up
  algorithm} (for a survey, see \cite{rachkovskij2012building}).  For
larger datasets, we hypothesize that these differences will be even
more pronounced.  Although each comparison by both MAC and {\sf
  Spontol-Retrieve} is a fast vector operation, for very large
datasets (e.g., $10^9$ relational structures), even a linear number of
vector operations becomes impractical.
In future work, we will test these systems on a broader range of
relational datasets to help elucidate the conditions under which
Spontol yields high accuracy and very-low retrieval cost.

\section{Conclusion} 
\label{section:conclusion}


The chief contribution of this paper is a demonstration of a system,
Spontol, that is able to solve the problem of spontaneous analogy.
That is, we have demonstrated how Spontol can efficiently store and
retrieve analogs without the need of human delineation of schemas.

Our representation also offers a new solution for the {\em binding
  problem} for long-term (static) memory that allows for efficient
analog retrieval in the absence of explicitly segmented domains.
The binding problem asks how we can meaningfully represent bindings
between roles and fillers.  Most solutions to the binding problem in
connectionism do so in terms of temporal synchronicity (e.g., LISA
\cite{hummel+holyoak:2005}).
Temporal synchronicity only works for knowledge in {\em working}
memory, and these models typically address storage in long-term memory
by relying on some form of conjunctive coding or tensor products.
Though these systems fail to address how relational structures can be
efficiently retrieved from long-term memory, we hypothesize
that a working-memory system, such as LISA, is necessary for the
``chaining'' process on which our system relies.


Spontol may offer evidence in support of a uniform
``substrate'' of intelligence \cite{mountcastle1978organizingShort}.  In
particular, we've shown how a system that was designed to process
perceptual data (Ontol) can be leveraged to process ``symbolic'' data
(i.e., relational structures).
This may provide insight into how species capable of higher-order
cognition might have evolved from species capable of only low-level
perception.

Although Spontol addresses some outstanding problems in Computational
Analogy, there is still ample room for future work.
Our implementation for characterizing a relational structure as a set
of windows might not scale well to very large structures without some
modifications.  An open problem is how windows might be managed in a
sensible way.  Spontol currently uses ``bags of windows'' for
medium-sized structures.  We propose extending Spontol by allowing
hierarchies of progressively higher-order bags to represent larger
structures (e.g., bags of bags of bags of windows).

\bibliographystyle{apacite}
\setlength{\bibleftmargin}{.125in}
\setlength{\bibindent}{-\bibleftmargin}
\bibliography{marcbib}

\end{document}